\documentclass{article} 
\usepackage{nips13submit_e,times}
\usepackage[numbers]{natbib}
\usepackage{hyperref}
\usepackage{url}
\usepackage{graphicx}
\usepackage{amsmath}
\usepackage{amsfonts}
\usepackage{subfigure}
\usepackage{multicol}
\usepackage[font=small]{caption}
\newtheorem{theorem}{Theorem}

\title{Multimodal Transitions for Generative Stochastic Networks}

\author{
Sherjil Ozair, Li Yao, \&  Yoshua Bengio\\
D\'epartement d'informatique et de recherche op\'erationnelle\\
Universit\'e de Montr\'eal\\
Montr\'eal, QC H3C 3J7 \\
}

\nipsfinalcopy 

\begin{document}

\maketitle

\begin{abstract}
Generative Stochastic Networks (GSNs) have been recently introduced
as an alternative to traditional probabilistic modeling: instead of
parametrizing the data distribution directly, one parametrizes a transition 
operator for a Markov chain whose stationary distribution is an estimator
of the data generating distribution. The result of training is therefore
a machine that generates samples through this Markov chain. However,
the previously introduced GSN consistency theorems suggest that in order
to capture a wide class of distributions, the transition operator in
general should be multimodal, something that has not been done before
this paper. We introduce for the first time {\em multimodal transition distributions} for GSNs,
in particular using models in the NADE
family ({\em Neural Autoregressive Density Estimator}) as output distributions
of the transition operator.
A NADE model is related to an RBM (and can thus model multimodal distributions)
but its likelihood (and likelihood gradient) can be computed easily.
The parameters of the NADE are obtained as a learned function of the previous
state of the learned Markov chain. Experiments clearly illustrate the advantage
of such multimodal transition distributions over unimodal GSNs.
\end{abstract}

\section{Introduction}

A new approach to unsupervised learning of probabilistic models has
recently been proposed~\citep{Bengio-et-al-NIPS2013-small,Bengio+Laufer-arxiv-2013}, 
called {\em Generative Stochastic Networks (GSNs)}, based
on learning the operator (transition distribution) of a Markov chain that
generates samples from the learned distribution. In Section~\ref{sec:gsn}, GSNs are formally defined and mathematical
results are provided on the consistency achieved when training GSNs with 
a denoising criterion. Denoising had previously been proposed as a criterion for unsupervised feature
learning of {\em denoising 
auto-encoders}~\citep{VincentPLarochelleH2008-small,Vincent-JMLR-2010-small}, one of the
successful building blocks for {\em deep learning}~\citep{Bengio-2009-book}.
The motivation for GSNs is that
the transition distribution of a Markov chain, given the previous state, is generally
simpler than the stationary distribution of the Markov chain, i.e., it has a
simpler structure and less major modes (the stationary distribution can be viewed
as a huge mixture, over all states, of these transition distributions). This could make it easier to
learn GSNs because one factor in the difficulty of training probabilistic
models is the complexity of approximating their normalizing constant
(partition function). If a distribution has fewer major modes or less complicated structure in them
(in the extreme,
if it is factorized or unimodal), then its partition function can be computed
or approximated more easily. However, previous work with GSNs has focused
only on the extreme case where the transition distribution is parametrized
by a factorized or unimodal distribution. This paper starts by reminding
the reader of good reasons for unsupervised learning of generative
models (Section~\ref{sec:unsupervised}) and two of the fundamental challenges
involved, namely the difficulty of mixing between modes and presence
of a potentially very large number of non-tiny modes. It argues in favor of
GSNs with multi-modal transition distributions (Section~\ref{sec:multimodal}) to address the
second problem and proposes (Section~\ref{sec:NADE}) to use conditional NADE~\citep{Larochelle+Murray-2011,Boulanger+al-ICML2012-small}
models (similar to conditional RBMs~\citep{Taylor+2007,TaylorHintonICML2009} 
but with a tractable likelihood) to represent multi-modal
output distributions for these transition operators.
Experiments are performed with denoising auto-encoder neural networks,
which were originally proposed as simple implementations of GSNs (Section~\ref{sec:exp}).
The results on both artificial (2D for easy visualization) and real data (MNIST)
clearly show that multimodality helps GSNs to better capture
the data generating distribution, reducing spurious modes by allowing
the transition operator to not be forced to put probability mass in 
the middle of two or more major modes
of the true transition distribution.

\section{Promises and Challenges of Unsupervised Learning of Generative Models}
\label{sec:unsupervised}

Unsupervised learning remains one of the core challenges for progress in
deep learning algorithms, with current applications of deep learning mostly
based on supervised deep nets. Unsupervised learning could hold the
key to many advantages for which purely supervised deep learning is
inadequate: 
\begin{itemize}
\item It allows a learner to take advantage of unlabeled data. Most of the
data available to machines (and to humans and animals) is unlabeled, i.e.,
without a precise and symbolic characterization of its semantics and of the
outputs desired from a learner.
\item It allows a learner to capture enough information about the
observed variables so as to be able to answer {\em new questions} about
them in the future, that were not anticipated at the time of seeing the
training examples.
\item It has been shown to be a good regularizer for supervised 
learning~\citep{Erhan+al-2010}, meaning that it can help the learner
generalize better, especially when the number of labeled examples is
small. This advantage clearly shows up (e.g., as illustrated in the transfer learning competitions
with unsupervised deep 
learning~\cite{UTLC+DL+tutorial-2011-small,UTLC+LISA-2011-small,Goodfellow+all-NIPS2011})
in practical applications where
the distribution changes or new classes or domains are considered
(transfer learning, domain adaptation), when some classes are frequent while
many others are rare (fat tail or Zipf distribution), or when new classes are shown
with zero, one or very few examples (zero-shot and one-shot learning ~\citep{Larochelle2008,Lake-et-al-NIPS2013}).
\item There is evidence suggesting that unsupervised learning can be
successfully achieved mostly from a {\em local training signal}
(as indicated by the success of the unsupervised layer-wise
pre-training procedures~\citep{Bengio-2009-book}, semi-supervised
embedding~\citep{WestonJ2008-small}, and intermediate-level hints~\citep{Gulcehre+Bengio-ICLR2013}), i.e.,
that it may suffer less from the difficulty of propagating credit
across a large network, which has been observed for supervised
learning. 
\item A recent trend in machine learning research is to apply
machine learning to problems where the output variable is very
high-dimensional (structured output models), instead of just
a few numbers or classes. For example, the output could be
a sentence or an image. In that case, the mathematical and
computational issues involved in unsupervised learning also arise.
\end{itemize}

What is holding us back from fully taking advantage of all these advanges?
We hypothesise that this is mainly because
unsupervised learning procedures for flexible and expressive models based
on probability theory and the maximum likelihood principle or its close
approximations \citep{MurphyBook2012} are almost all facing the {\em challenge of marginalizing}
(summing, sampling, or maximizing) {\em across many explanations} or {\em across many
  configurations of random variables}.  This is also true of structured
output models, or anytime we care about capturing complicated joint
probabilistic relationships between many variables.  This is best
exemplified with the most classical of the deep learning algorithms, i.e.,
Boltzmann machines~\citep{Ackley85} with {\em latent variables}.
 In such models, one faces the two common
intractable sums involved when trying to compute the data likelihood or its
gradient (which is necessary for learning), or simply to use the trained
model for new inputs:
\begin{itemize}
\item The intractable sum or maximization
over latent variables, given inputs, which is necessary for {\em inference},
i.e., when using the model to answer questions about new examples,
or as part of the training procedure.
\item The intractable sum or maximization over all the variables (the observed
ones and the latent ones), which is necessary for {\em learning}, in
order to compute or estimate the gradient of the log-likelihood, in
particular the part due to the normalization constant of the
probability distribution. This intractable
sum is often approximated with Monte-Carlo Markov Chain (MCMC) methods,
variational methods, or optimization methods (the MAP or Maximum 
A Posteriori approximation).\footnote{The partition function problem
goes away in directed graphical models, shifting completely the burden
to inference.}
\end{itemize}
In principle, MCMC methods should be able to handle both of these
challenges. However, MCMC methods can suffer from two fundamental issues,
discussed in~\citet{Bengio+Laufer-arxiv-2013} in more detail:
\begin{itemize}
\item {\bf Mixing between well-separated modes.} A mode is a local maximum
  of probability (by extension, we refer to a connected region of high
  probability). According to the {\em manifold hypothesis}\footnote{stating
    that probability is concentrated in some low-dimensional regions;
    the manifold hypothesis holds for most
    AI tasks of interest because most configurations of input variables are
    unlikely.}~\citep{Cayton-2005,Narayanan+Mitter-NIPS2010-short},
  high-probability modes are separated by vast areas of low
  probability. Unfortunately, this makes it exponentially difficult for
  MCMC methods to travel between such modes, making the resulting
  Monte-Carlo averages unreliable (i.e., of very high variance).
\item {\bf A potentially huge number of large modes.} It may be
the case that the number of high-probability modes is very large (up to exponentially large),
when considering problems with many explanatory factors. When this is so,
the traditional approximations based on MCMC, variational methods,
or MAP approximations could all fall on their face. In particular, since
MCMC methods can only visit one mode at a time, the resulting estimates
could be highly unreliable.
\end{itemize}
These are of course hypotheses, they need to be better tested, and may be
more or less severe depending on the type of data generating distribution
being estimated. The work on GSNs aims at giving us a tool to test these
hypotheses by providing algorithms where the issue of normalization constant
can be greatly reduced.

\section{Generative Stochastic Networks}
\label{sec:gsn}

A very recent result~\citep{Bengio-et-al-NIPS2013-small,Bengio+Laufer-arxiv-2013}
shows both in theory and in
experiments that it is possible to capture the data generating distribution
with a training principle that is completely different from the maximum
likelihood principle.  Instead of directly modeling $P(x)$ for observed
variables $x$, one learns the transition operator of a Markov chain whose stationary distribution
estimates $P(x)$. The idea is that a conditional distribution of a Markov chain
typically makes only ``small'' or ``local'' moves\footnote{if it did not make a move sampled
from a distribution with small
entropy it would have to reject most of the time, according to the manifold
hypothesis stating that probability mass is concentrated}, 
making the {\em conditional distribution} simpler in the
sense of having {\em fewer dominant modes}. The surprising result is that there
is at least one simple training criterion for such a transition operator, based on
probabilistically undoing the perturbations introduced by some noise source.
The transition operator of the Markov
chain, $P(x_t, h_t | x_{t-1}, h_{t-1})$ (with observed variable $x$ and
latent state variables $h$)
can be trained by using a {\em denoising} objective, in which $x_{t-1}=x$
is first mapped into a learned $f(x_{t-1},h_{t-1},z_t)$ (with a noise source $z_t$)
that destroys information about $x$,
and from which the clean data point $x$ is predicted and its
probability $P(x_t=x | f(x_{t-1}=x,h_{t-1},z_t))$ approximately maximized.
Since the corruption only leaves a relatively small number of
configurations of $x$ as probable explanations for the computed
value of $f(x_{t-1}=x,h_{t-1},z_t)$, the reconstruction
distribution $P(x_t=x | f(x_{t-1}=x,h_{t-1},z_t))$ can generally be approximated
with a much simpler model (compared to $P(x)$). In a sense, this is something
that we are directly testing in this paper.

The GSN framework therefore only addresses the issue of a potentially huge
number of modes (problem 2, above). However, by training GSNs with deep
representations, one can hope to take advantage of the recently observed
superiority of trained representations (such as in deep auto-encoders or
deep belief nets and deep Boltzmann machines) in terms of faster mixing between
well separated modes~\citep{Bengio-et-al-ICML2013}.

The core mathematical result justifying this procedure is the following theorem,
proven in~\citet{Bengio+Laufer-arxiv-2013}, and stating sufficient conditions
for the denoising criterion to yield a consistent estimator of the data generating
distribution, as the stationary distribution of the Markov chain:
\begin{theorem}
\label{thm:gsn}
Let training data $X \sim {\cal P}(X)$ and independent noise $Z \sim {\cal P}(Z)$ and 
introduce a sequence of latent variables $H$ defined iteratively through a function 
$f$ with
$H_t = f_{\theta_1}(X_{t-1},Z_{t-1},H_{t-1})$ for a given sequence of $X_t$'s.
Consider a model $P_{\theta_2}(X | f_{\theta_1}(X, Z_{t-1},H_{t-1}))$ trained
(over both $\theta_1$ and $\theta_2$) so that $P_{\theta_2}(X | H)$,
for a given $\theta_1$, is a consistent estimator of the true
${\cal P}(X|H)$.
Consider the Markov chain defined above and assume that it converges to a stationary 
distribution $\pi_n$ over the $X$ and $H$ and with marginal $\pi_n(X)$, even in the limit
as the number of training examples 
$n \rightarrow \infty$.  
Then $\pi_n(X) \rightarrow {\cal P}(X)$ as $n \rightarrow \infty$.
\end{theorem}

A particularly simple family of GSNs is the denoising auto-encoder,
studied in~\cite{Bengio-et-al-NIPS2013-small} as a GSN, i.e., as a generative model.
In that case, $H_t=f_{\theta_1}(X_{t-1}, Z_{t-1},H_{t-1})$ is just the parameter-less
corruption of $X_{t-1}$ according to a predetermined corruption process such as
adding Gaussian noise or multiplying by masking noise, also known
as dropout (and $H_t$ does not depend on a previous $H_{t-1}$).

In ~\citet{Bengio+Laufer-arxiv-2013}, this
approach is used to successfully train an unsupervised architecture
mimicking the structure of a Deep Boltzmann Machine. In~\citet{Goodfellow-et-al-NIPS2013}
the same principle is used to provide a successful sampling scheme for the
Multi-Prediction Deep Boltzmann Machine (MP-DBM), when its training procedure is viewed as a particular
form of GSN. Note that the MP-DBM was shown to be particularly successful at classification
tasks, in spite of being trained with both the image and class being treated as visible variables
in a single integrated training framework, unlike previous approaches for training DBMs~\citep{Salakhutdinov2009}.
Much more work
is needed to better understand the GSN approach and expand it to
larger models, structured output problems, and applications, and more importantly,
to explore the space of architectures and models that can be trained
by this novel principle. In this paper, we focus on one particular
aspect of research on GSNs, namely the need for multimodality of the
{\em transition distribution}, the one that computes the probability of the next
visible state given the previous state, $P(x_t | f(x_{t-1},h_{t-1},z_t))$.

\section{The Need for Multimodality of the Transition Distribution}
\label{sec:multimodal}

Close inspection of Theorem~\ref{thm:gsn} reveals that if the ``true''
data generating distribution requires $P(x_t=x | f(x_{t-1}=x,h_{t-1},z_t))$ 
to be multimodal, then that capability is required of our parametrization
of the transition probability in order to get consistency. Otherwise we
only get consistency in the family of functions that can be represented
by such unimodal transition operators. We already know from a mathematical
result in~\citet{Alain+Bengio-ICLR2013} that
{\em when the amount of corruption noise converges to 0 and the input variables
have a smooth continuous density, then a unimodal Gaussian reconstruction
density suffices to fully capture the joint distribution}. The price
to pay for this easier parametrization of the conditional distribution
is that the associated Markov chain would mix very slowly, making it much
less useful as a generative model. At the other extreme, if the amount
of corruption noise is ``infinite'' (destroying all information
about the previous state), then $P(x_t=x | f(x_{t-1}=x,h_{t-1},z_t))=P(x_t=x)$
and we are back to a normal probabilistic model, which generally has a lot
of modes. 

\begin{figure}[ht]
\centering
\includegraphics[width=0.8\textwidth]{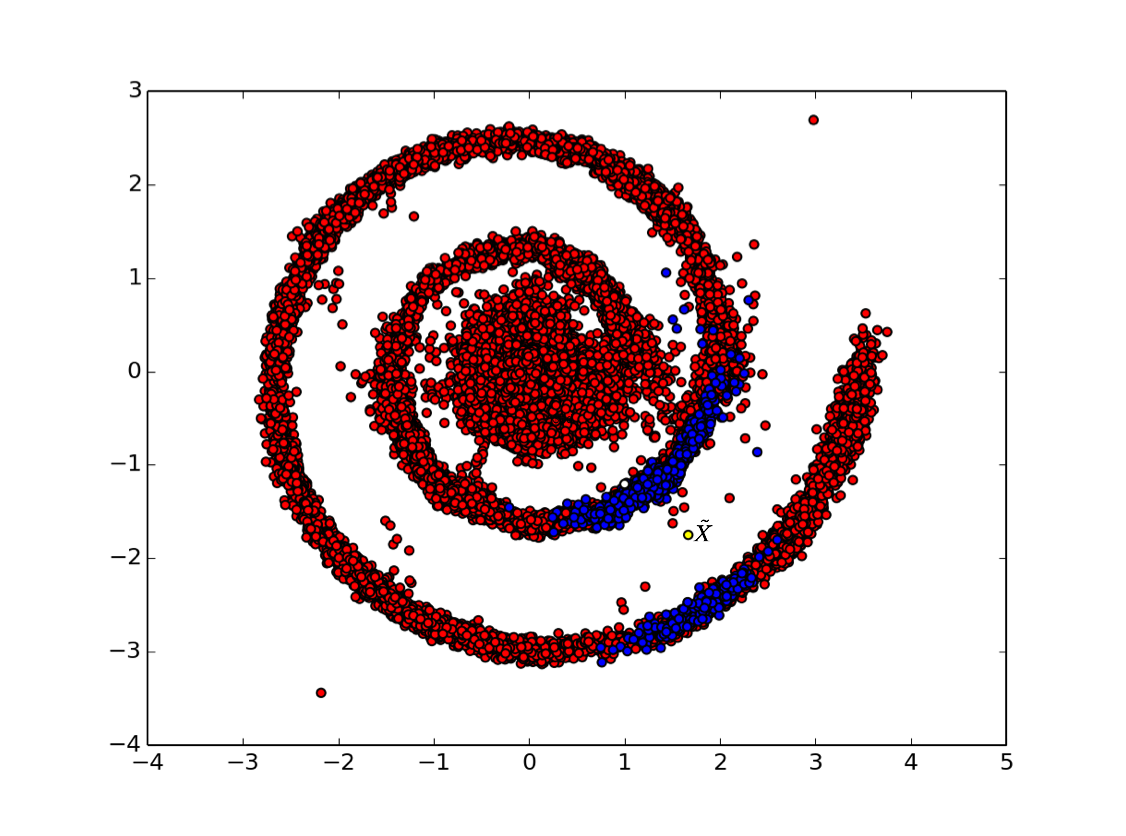}
\caption{
Example illustrating the need for a multimodal reconstruction distribution. 
The red dots are samples from the model ($P(X)$), the yellow dot
is a sample from the latent variable (the corrupted sample $\tilde{X}$
of a denoising auto-encoder, obtained by adding isotropic Gaussian noise
with standard deviation 0.3) and the blue dots show samples of the
multimodal reconstruction distribution $P(X|\tilde{X})$. Clearly, there
are two main modes from which $X$ could have come to produce $\tilde{X}$.
A unimodal (e.g., factorial) distribution for $P(X|\tilde{X})$ could
not capture that and one would obtain as estimator a wide Gaussian covering
both modes, i.e., corresponding to generating a lot of spurious samples 
in between the real modes.
}
\label{fig:multimodal-example}
\end{figure}

The advantage of multimodal GSNs such as discussed in this paper is that
they allow us to explore the realm in between these extremes, where the
transition operator is multimodal but yet the number of such modes remains
small, making it relatively easy to learn the required conditional
distributions. This is illustrated in Figure~\ref{fig:multimodal-example},
which shows an example where having a unimodal reconstruction distribution
(factorial GSN, i.e., a regular denoising auto-encoder with factorized Gaussian 
or factorized Bernoulli output units) would yield a poor model of the data, with spurious samples generated in
between the arms of the spiral near which the data generating distribution
concentrates.

\section{NADE and GSN-NADE Models}
\label{sec:NADE}

The {\em Neural Auto-regressive Density Estimator} (NADE) family of probability models~\citep{Larochelle+Murray-2011}
can capture complex distributions (more so by increasing capacity, just like other
neural nets or mixture models) while allowing the analytic and tractable computation
of the likelihood. It is a descendant of previous neural ``auto-regressive''
neural networks~\citep{Bengio+Bengio-NIPS99}, also based on decomposing $P(x_1, x_2, \ldots x_n)$
into $\prod_i P(x_i | x_1, \ldots x_{i-1})$, with a group of output units
for each variable $i$ and an architecture that prevents input $x_j$ to enter into
the computation of $P(x_i | x_1, \ldots x_{i-1})$ when $j\geq i$.
Whereas the models of~\citet{Bengio+Bengio-NIPS99,Larochelle+Murray-2011} were developed for discrete
variables, NADE was recently extended~\citep{Benigno-et-al-NIPS2013-small} 
to continuous variables by making $P(x_i | x_1, \ldots x_{i-1})$
a mixture of $k$ Gaussians (whose means and variances may depend on $x_1, \ldots x_{i-1}$).

Like other parametric models, NADE can be used as the output distribution of a conditional
probability. This has already been done in the context of modeling musical sequences~\citep{Boulanger+al-ICML2012-small},
with the outputs of a recurrent neural networks being the bias parameters of a NADE model
capturing the conditional distribution of the next frame (a set of musical notes), given
previous frames (as captured by the state of the recurrent neural network). In this paper
we propose to use a NADE or RNADE model to capture the output distribution of a transition
operator for the Markov chain associated with a GSN. More specifically, the experiments
performed here are for the case of a denoising auto-encoder, i.e., a neural network takes
the corrupted previous state $\tilde{X}$ as input and attempts to denoise it probabilistically,
i.e., it outputs the parameters of a NADE model (here, again the hidden and output layer biases)
associated with the distribution of the next state (i.e. the clean input $X$), given $\tilde{X}$.
The weights of the NADE model are kept unconditional: because these weight matrices can be large, it
makes more sense to try to only condition the biases, like has been done before with
conditional RBMs~\citep{Taylor+2007,TaylorHintonICML2009}.

\section{Experiments}
\label{sec:exp}

\begin{figure}[ht]
\begin{center}
\includegraphics[width=0.45\textwidth]{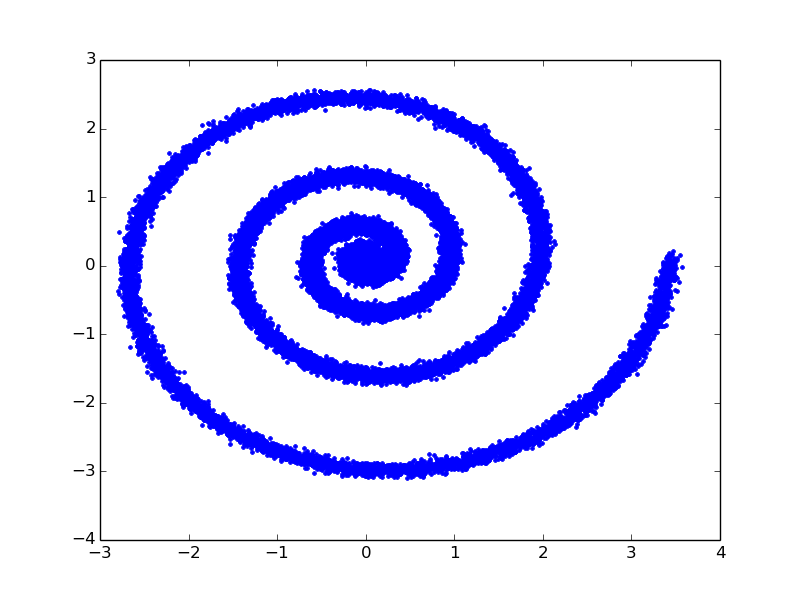}
\end{center}
\includegraphics[width=0.45\textwidth]{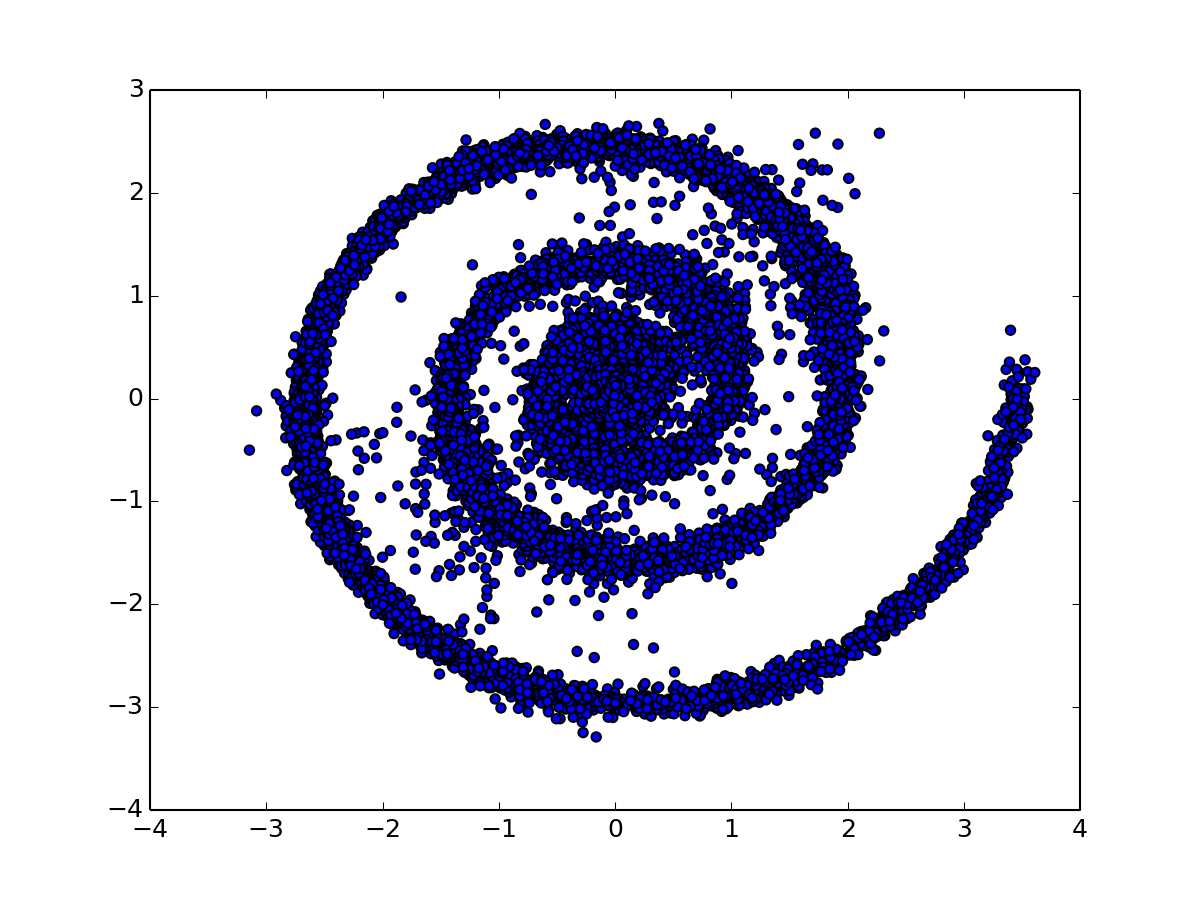}
\includegraphics[width=0.45\textwidth]{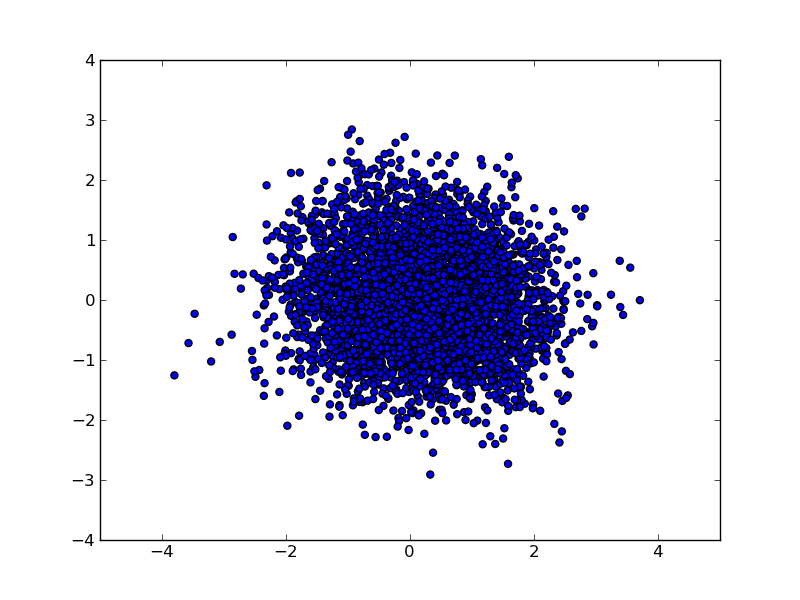}
\caption{
Top: Training examples. Bottom left: Samples from a GSN-NADE model with 5 Gaussian components per NADE
output unit. Bottom right: samples from a GSN model with factorized reconstruction distribution. 
}
\label{fig:multimodal-example}
\end{figure}

In order to demonstrate the advantages of using a multimodal reconstruction distribution 
over the unimodal one, we firstly report experimental results on a 2D real-valued dataset. 
Two types of GSNs are trained and then used to generate samples, shown in Figure \ref{fig:multimodal-example}.
Both GSN-NADE and fatorial GSN use Gaussian corruption $p(\tilde{X}|X)=X + \mathcal{N}(0, \sigma)$. To make 
the comparison fair, both models use $\sigma=0.3$. When the added noise is significantly large, using a 
multimodal reconstruction distribution $p(X|\tilde{X})$ is particularly advantageous, as is clearly 
demonstrated in the Figure. The samples generated by the GSN-NADE model resemble closely to the original 
training examples.

\begin{figure}[ht]
\begin{minipage}{1.0\linewidth}
\centering
\includegraphics[width=0.6\textwidth]{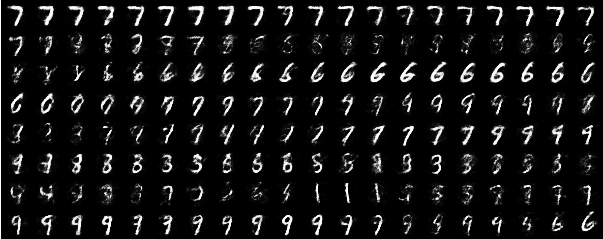} 
\caption{Consecutive samples generated by training GSN-1, the single hidden layer GSN 
with no walkbacks.}
\label{mnist:dae}
\end{minipage}

\begin{minipage}{1.0\linewidth}
\centering
\includegraphics[width=0.6\textwidth]{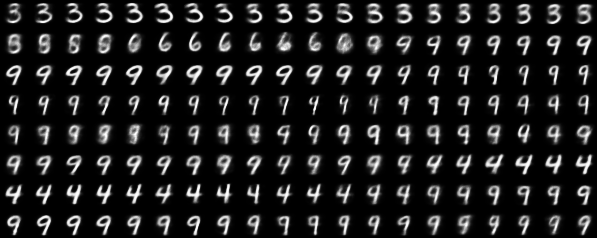} 
\caption{Consecutive samples generated by training GSN-1-w, the single hidden layer GSN 
with walkbacks.}
\label{mnist:gsn-walkback}
\end{minipage}
\begin{minipage}{1.0\linewidth}
\centering
\includegraphics[width=0.6\textwidth]{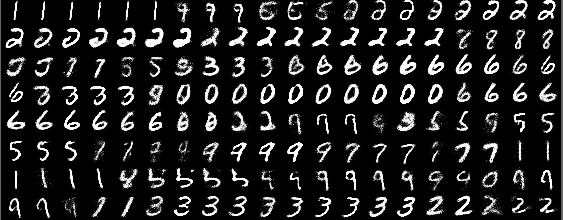}
\caption{Consecutive samples generated by training GSN-NADE with no walkback training.}
\label{mnist:gsn-nade}
\end{minipage}
\end{figure}

The second set of experiments are conducted on MNIST, the handwritten digits dataset.  
The real-valued pixels are binarized with the threshold 0.5. A 784-2000-2000-784 NADE model is 
chosen as the reconstruction distribution of the GSN-NADE model. The biases of the first 2000 hidden units 
of the NADE network are outputs of a 784-2000 neural network, which has $\tilde{X}$ as the input. The training of model uses the same procedure as in \citep{Benigno-et-al-arxiv-2013}. In addition, a dynamic noise is added on input pixels: each training example is corrupted with a salt-and-pepper noise that is uniformly sampled between 0 and 1. We find in practice that using dynamic noise removes more spurious modes in the samples. 
To evaluate the quality of trained GSNs as generative models, we adopt the Conservative 
Sampling-based Log-likelihood (CSL) estimator proposed in ~\citep{Bengio+Yao-arxiv-2013}. 
The CSL estimates are computed on different numbers of consecutive samples, all starting 
from a random intialization, as opposed to computing the estimates based on samples 
collected every 100 steps originally reported in ~\citep{Bengio+Yao-arxiv-2013}.
 The detailed comparison is shown 
in Table \ref{tab:CSL}. In the table, GSN-NADE denotes the model we propose in this work. 
The GSN-1-w is a GSN with 
factorial reconstruction distribution trained with the walkback procedure proposed 
in \citet{Bengio-et-al-NIPS2013-small}. 
GSN-1 denotes a GSN with a factorial reconstruction distribution with no walkback training. 
For the details of models being compared, please refer to 
~\citep{Bengio+Yao-arxiv-2013}. Figure \ref{mnist:dae}, Figure \ref{mnist:gsn-walkback} 
and Figure \ref{mnist:gsn-nade} show collections of consecutively 
generated samples from all 3 types of models after training.

\begin{table}[ht]
\centering
\caption{The CSL estimates 
    obtained using different numbers of {\em consecutive} 
    samples of latent variables: $H$ 
are collected with consecutive sampling steps of one Markov chain.}
\label{tab:CSL}
\begin{tabular}{lrrrrr}
\hline
\# samples & GSN-NADE & GSN-1  & GSN-1-w \\
\hline
10k        &-143    & -148  & -127   \\
50k        &-124    & -131  & -109   \\
100k       &-118    & -125   & -107   \\
150k       &-114    & -121   & -106   \\
\hline
\end{tabular}
\end{table}
 
Experiments confirm that the GSN model with a NADE output
distribution does a much better job than the GSN model with a factorized
output distribution on learning and generating samples for the real-valued 2 dimensional spiral
manifold (Figure \ref{fig:multimodal-example}). On MNIST, the differences between generated samples are also visually striking. 
As demonstrated in the picture, 
a factorial GSN without the help of the walkback training (Figure \ref{mnist:dae}) tends to generate 
sharper digits but suffers from often getting stuck in the non-digit-like spurious modes. 
On the contrary, a GSN-NADE without the walkback training (Figure \ref{mnist:gsn-nade}) alleviates this issue. In addition, GSN-NADE mixes much better among digits, 
It generates samples that are visually much better than the rest two factorial models. To show that GSN-NADE mixes better, we also computed the CSL estimate of the same 
GSN-NADE reported in Table \ref{tab:CSL} with 10k samples but all of which are collected every 100 steps. 
Surprisingly, the CSL estimate remains exactly the same as the one where samples are collected 
after every step of the Markov chain (shown in Table \ref{tab:CSL}).  
Furthermore, as shown in Figure \ref{mnist:gsn-nade}, its samples present more diversity in writing styles, 
directly resulting a significantly better CSL estimate (CSL estimates are 
in log scale) 
than its factorial counterpart. GSN-NADE, however, is not able to outperform the factorial GSN trained with the walkback 
training (Figure \ref{mnist:gsn-walkback}) in terms of the CSL estimates. This is because factorial GSNs trained with walkbacks win by suppressing almost all the spurious modes, resulting 
in higher CSL estimates of the testset log-likelihood (CSL prefers blury digits to spurious modes). 

\section{Conclusion}

In this paper we have motivated a form of generative stochastic networks (GSNs) in which the
next-state distribution is multimodal and we have proposed to use conditional NADE distributions to capture
this multimodality. Experiments confirm that this allows the GSN to more easily capture the
data distribution, avoiding the spurious samples that would otherwise be generated by a unimodal
(factorial) output distribution (the kind that has been used in the past for GSNs and denoising
auto-encoders).

In the MNIST experiments, we found that the benefit of the walkback procedure was actually
greater than the benefit of the multimodality brought by the NADE output distribution.
How about combining both NADE and walkback training?
The basic procedure for sampling from NADE is expensive because it involves sampling each input variable
given all the previous ones, each time having to run through the NADE neural net.
Applying the walkback training procedure to the GSN-NADE models
therefore poses a computational problem because the walkback procedure involves sampling
a few steps from the chain in the middle of the training loop (in a way similar to contrastive
divergence training): whereas getting the likelihood gradient is
fast with NADE (order of number of inputs times number of hidden units),
sampling from NADE is much more expensive (multiply that by the number
of inputs). Future work should therefore investigate other multimodal distributions
(for which sampling is cheap) or ways to approximate the NADE sampling procedure
with faster procedures.

\subsubsection*{Acknowledgments}
We would like to thank the developers of Theano~\citep{bergstra+al:2010-scipy,Bastien-Theano-2012}
and Pylearn2~\citep{pylearn2_arxiv_2013}. We would also like to thank NSERC, the Canada Research
Chairs, and CIFAR for funding, as well as Compute Canada, and Calcul Qu\'ebec
for providing computational resources.
\bibliography{strings,strings-shorter,ml,aigaion}
\bibliographystyle{natbib}
\end{document}